\pdfoutput=1

\documentclass[runningheads]{llncs}

\usepackage{lmodern}
\usepackage[T1]{fontenc}
\usepackage[utf8]{inputenc}
\usepackage{textcomp}
\usepackage{microtype}
\usepackage{hyphenat}
\usepackage[all]{nowidow}
\usepackage{enumitem}
\usepackage[normalem]{ulem}

\usepackage[
    width=122mm,
    left=12mm,
    paperwidth=146mm,
    height=193mm,
    top=12mm,
    paperheight=217mm]{geometry}

\usepackage[
    hidelinks=true,
    breaklinks=true,
    letterpaper=true,
    bookmarks=false]{hyperref}

\usepackage{amssymb}
\usepackage{amsmath}

\usepackage{graphicx}
\usepackage{float}
\usepackage{subcaption}
\usepackage[labelsep=period,labelfont=bf,font={small}]{caption}

\usepackage[square,numbers]{natbib}
\usepackage{booktabs}

\usepackage{xcolor}
\usepackage{soul}

\hypersetup{pdfcreator={Rijnder Wever, Tom F.H. Runia}}
\hypersetup{pdfproducer={Rijnder Wever, Tom F.H. Runia}}

\hypersetup{pdftitle={Subitizing with Variational Autoencoders}}
\hypersetup{pdfsubject={Computer Vision}}
\hypersetup{pdfkeywords={vision, counting, numerosity, subitizing, object detection}}

\title{Subitizing with Variational Autoencoders}

\definecolor{highlightcolor}{RGB}{255,237,191}
\definecolor{textboxcolor}{RGB}{155,155,155}
\definecolor{paperboxcolor}{RGB}{204,148,73}
\definecolor{highlightred}{RGB}{224,150,150}
\definecolor{highlightgreen}{RGB}{150,224,150}

\begin{document}

\pagestyle{headings}
\mainmatter
\def\ECCV18SubNumber{W64P20}

\title{Subitizing with Variational Autoencoders} 
\titlerunning{Subitizing with Variational Autoencoders}


\authorrunning{Rijnder Wever, Tom F.H. Runia}
\author{Rijnder Wever, Tom F.H. Runia}
\institute{University of Amsterdam, Intelligent Sensory Information Systems}

\maketitle

\begin{abstract}
Numerosity, the number of objects in a set, is a basic property of a given visual scene. Many animals develop the perceptual ability to \emph{subitize}: the near-instantaneous identification of the numerosity in small sets of visual items. In computer vision, it has been shown that numerosity emerges as a statistical property in neural networks during unsupervised learning from simple synthetic images. In this work, we focus on more complex natural images using unsupervised hierarchical neural networks. Specifically, we show that \emph{variational autoencoders} are able to spontaneously perform subitizing after training without supervision on a large amount images from the Salient Object Subitizing dataset. While our method is unable to outperform supervised convolutional networks for subitizing, we observe that the networks learn to encode numerosity as basic visual property. Moreover, we find that the learned representations are likely invariant to object area; an observation in alignment with studies on biological neural networks in cognitive neuroscience.
\end{abstract}

\vspace*{-0.76cm}

\section{Introduction}\label{introduction}

The ability to answer the question ``How many?'' is an important capability of our visual recognition system. Animals use \emph{visual number sense} to rank, quantify and label objects in a scene \cite{nieder2009representation}. There is evidence \cite{nieder2016neuronal, piazza2009humans,burr2008visual} that the human brain uses at least two distinct mechanisms for non-verbal representation of number: one for large quantity \emph{estimation} and a \emph{subitizing} faculty for near instantaneous identification of a small number of objects ($1$-$4$). In this work, we propose a brain-inspired approach for learning to subitize from large image datasets.

The concepts of visual number sense and instance counting are well studied in computer vision \cite{lempitsky2010learning,segui2015learning,zhang2016salient,liu2016highway,arteta2016counting,walach2016learning,xiong2017spatiotemporal,he2017delving,Chattopadhyay_2017_CVPR}. Most recent work relies on supervised learning using deep convolutional neural networks (CNNs) \cite{lecun2015deep} to assess the instance count in a given scene. While existing methods perform admirably on various counting tasks, the number of visual classes and the availability of labeled image datasets is decisive for their performance. Motivated by these observations and our visual cognition system, this paper explores hierarchical representation learning for visual numerosity in an \emph{unsupervised} setting. 

\begin{figure}[t]
\centering
    \begin{subfigure}{.19\linewidth}
        \includegraphics[width=\textwidth]{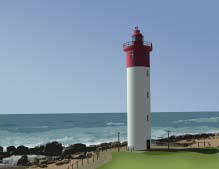}
    \end{subfigure}
    \begin{subfigure}{.19\linewidth}
        \includegraphics[width=\textwidth]{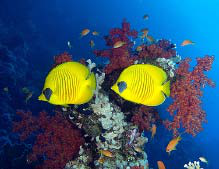}
    \end{subfigure}
    \begin{subfigure}{.19\linewidth}
        \includegraphics[width=\textwidth]{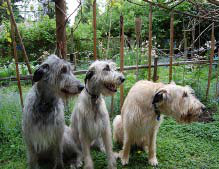}
    \end{subfigure}
    \begin{subfigure}{.19\linewidth}
        \includegraphics[width=\textwidth]{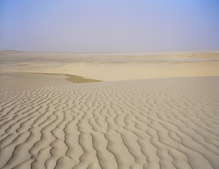}
    \end{subfigure}
    \begin{subfigure}{.19\linewidth}
        \includegraphics[width=\textwidth]{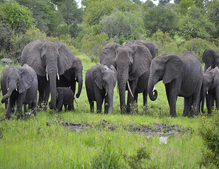}
    \end{subfigure}
    \caption{Example images from the Salient Object Subitizing dataset \cite{zhang2016salient}. Although the ability to \emph{subitize} should allow people to identify the number of instances in each image at a glance, these scenes pose a challenge to computer vision models due to variety in appearance, saliency ambiguity, scene clutter and occlusion.}
\label{fig:intro-figure}
\vspace{-1em}
\end{figure}

Our work is inspired by the observation of \citet{stoianov2012} that visual numerosity emerges as a statistical property of images in artificial neural networks trained in an unsupervised manner. Specifically, the authors train Restricted Boltzmann Machines (RBMs) on synthetic images containing a random number of objects and show that neural response distributions correlate with number discriminability. Their observations are intriguing but the simple synthetic images do not capture the complexity of natural visual scenes. In this work, we focus on unsupervised learning of numerosity representations from natural images containing diverse object classes (see example images in \autoref{fig:intro-figure}). 

The contributions of this work are the following. We explore the emergence of visual number sense in deep networks trained in an unsupervised setting on natural images. Specifically, we propose the use of variational autoencoders with both the encoder and decoder parametrized as CNNs to effectively handle complex images and maintain spatial organization. For optimization, we include the recently proposed feature perceptual loss \cite{hou2017deep} instead of the pixel-to-pixel distance metric to aid representation learning. Finally, we present preliminary quantitative and qualitative results on unsupervised representation learning for numerosity from the Salient Object Subitizing dataset \cite{zhang2016salient}. 

\section{Related Work}
\label{related-work}

\noindent\textbf{Numerical Cognition.} Non-verbal numerical competence is implicitly developed across humans \cite{lakoff, feigenson2004core, franka2008number, dehaene2011number} and animal species \cite{nieder2016neuronal, harvey2013topographic, animalsnumericalcognition}. These abilities possibly arise from numerosity being an integral part of the sensory world \cite{viswanathan2013neuronal}. Interestingly, humans have developed the ability to \emph{subitize} \cite{jevons1871power,kaufman1949discrimination} for near-instantaneous numerosity identification of small visual sets ($1$-$4$ items). The near-instantaneous character of subitizing and it's spontaneous neural development are possibly caused by the visual system's limited but automatic capability to process spatial configurations of salient objects \cite{piazza2009humans, jansen2014role, dehaene2011number}. Whereas the visual number sense relates to properties as object area and density, neural responses of numerosity-selective cognitive systems were found to be invariant to all visual features except quantity \citep{nieder2016neuronal,harvey2013topographic}. Furthermore, studies on cognitive neuroscience have shown that the perception of number functions independently from mathematical reasoning \cite{harvey2013topographic, poncet2016individuation}. All these findings suggest that visual number sense is a perceptual property that emerges directly from the visual sensory input. \\ 

\noindent\textbf{Numerosity in Computer Vision.} Instance counting in visual scenes has received substantial interest from the deep vision community, notably in object counting \cite{lempitsky2010learning,Chattopadhyay_2017_CVPR,he2017delving,liu2016highway,noroozi2017representation}, crowd-size estimation \cite{xiong2017spatiotemporal,hu2016dense}, animal population estimation \cite{arteta2016counting} and video repetition \cite{levy2015live,runia2018repetition}. The shared similarity between most of these recent works is their use of CNNs for supervised representation learning from large image datasets. Of these approaches, the recent work of \citet{zhang2016salient} is most similar to ours as we also evaluate on the task of instance counting and use their Salient Object Subitizing dataset. While these methods are effective on specific domains, they require large amounts of labeled data and are limited to a predefined set of visual classes and numerosity range. Therefore, we here study brain-inspired unsupervised representation learning for visual number sense. 

\citet{stoianov2012} discovered the emergence of neural populations in artificial neural networks sensitive to numerosity while invariant to object size. Their observations align with object size invariance for visual number sense in the human brain \cite{nieder2016neuronal}. In this work we emphasize on learning visual number representations from realistic natural images rather than the simple binary images studied in \cite{stoianov2012}. As a consequence, representation learning becomes significantly more challenging, making RBMs difficult to train. We propose variational autoencoders \cite{kingma2013auto} to learn visual numerosity representations in an unsupervised setting. 


\section{Methods}
\label{methods}

Inspired by \citet{stoianov2012} we propose an unsupervised generative model to learn visual numerosity representations from natural and synthetic image datasets. Specifically, we use a variational autoencoder for encoding and reconstructing training images. The underlying principle is that numerosity is a key characteristic in the images and the network learns to encode visual numerosity in the latent representation. 

\subsection{Variational Autoencoder}
\label{subsec:variational-autoenconder}

We use the original definition of the variational autoencoder (VAE) as introduced by \citet{kingma2013auto}. VAEs are among the most popular approaches for unsupervised representation learning due to their generative nature and the fact that the encoder and decoder can be parameterized by neural networks trainable with stochastic gradient descent. For an excellent overview of VAEs we refer the reader to the tutorial of \citet{doersch2016tutorial} as we here only outline the core idea.

VAEs learn to map data samples $X$ to a posterior distribution \mbox{$Q(z \mid X)$} rather than a deterministic latent representation as used in conventional autoencoders. Inputs can be reconstructed by sampling latent vector $z$ from the posterior distribution and passing it through a decoder network. To make sampling feasible, the posterior distribution is parametrized by a Gaussian distribution with its mean and variance predicted by the encoder. In addition to a reconstruction loss, the posterior $Q(z\mid X)$ is regularized with its Kullback–Leibler divergence from a prior distribution $P(z)$ which is typically also Gaussian with zero mean and unit variance such that the KL divergence can be computed in closed form \cite{kingma2013auto}. Together, the VAE's objective function is the summation of a reconstruction term (negative log-likelihood of the data) and the KL regularization:
\begin{align}
  \mathcal{L}_{VAE} = E[\log P(X \mid z)] - \mathcal{D}_{KL}[ Q(z \mid X) \mid\mid P(z)]
\end{align}
We use this formulation with both encoder and decoder parametrized as convolutional neural network to learn visual representations from a large collection of images displaying scenes with a varying number of salient objects. 

\begin{figure}[t]
    \centering
    \includegraphics[width=0.9\textwidth]{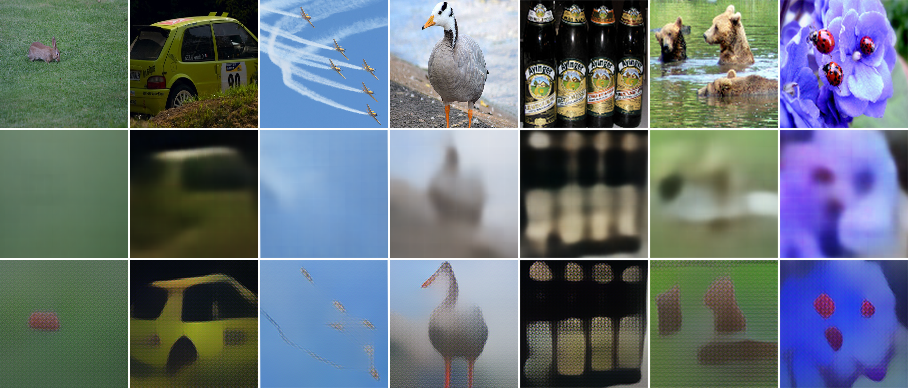}
    \caption{\textbf{Top:} original images from the Salient Object Subitizing dataset \cite{zhang2016salient}. \textbf{Center:}  VAE reconstructions using traditional loss. \textbf{Bottom:} VAE reconstructions using feature perceptual loss. Note the improved ability to reconstruct salient objects and contour sharpness, likely beneficial for object subitizing.}
    \label{fig:dfc_comp}
\end{figure}

\subsection{Feature Perceptual Loss}
\label{sec:feature-loss}

VAEs are known to produce blurry reconstructions \cite{goodfellow2016deep}. In our preliminary experiments we observed difficulties with reconstructing multiple salient objects, negatively affecting the ability to subitize. Therefore, we employ the recent feature perceptual loss of \citet{hou2017deep} which uses intermediate layer representations in the objective function of the autoencoder. The authors use a VGG-19 network \cite{simonyan2014very} pretrained on ImageNet \cite{russakovsky2015imagenet} denoted as $\Phi$ and define a set of layers $l_i \in L$ for computing the perceptual loss. Specifically, during the training the mean squared error between the hidden representations of input image $X$ and the reconstruction $\widetilde{X}$ is added to the loss for the predefined layers:
\begin{align}
    \mathcal{L}^{L}_{rec} = \sum_{l \in L}^{} \textrm{MSE}(\Phi(X)^{l}, \Phi(\widetilde{X})^{l}) \label{fpl}
\end{align}
The intuition is that responses of layers $l_i$ should be retained in reconstruction $\widetilde{x}$ as they represent important visual characteristics. Following their recommendations and our own findings, we use $L = \{\texttt{relu1\_1, relu2\_1, relu3\_1} \}$ from the pretrained VGG-19 network to compute the loss. The feature perceptual loss and original VAE objective are weighed according to $\mathcal{L}_{total} = \alpha\mathcal{L}_{kl} + \beta\mathcal{L}_{rec} \label{total}$ in which $\alpha$ and $\beta$ are hyperparameters. We found this extension to our autoencoder to improve visual saliency and representation learning compared to plain pixel-by-pixel reconstruction loss (see \autoref{fig:dfc_comp} for a visual comparison). 


\section{Experiments}
\label{sec:experiments}


\subsection{Datasets}
\label{subsec:datasets}

\noindent \textbf{Salient Object Subitizing Dataset.} Proposed by \citet{zhang2016salient}, the Salient Object Subitizing (SOS) dataset contains $14$K images for the purpose of instance counting. The images originate from MS-COCO \cite{lin2014microsoft}, ImageNet \cite{russakovsky2015imagenet} and SUN \cite{xiao2010sun}. Each image is annotated with an instance count label: $0$, $1$, $2$, $3$ or $4+$ salient objects. The final image collection is biased towards centered dominant objects and backgrounds scenes as the authors observe \cite{zhang2016salient}. In practice, the class imbalance may pose training difficulties. See \autoref{fig:intro-figure} for some examples.\\

\noindent \textbf{Synthetic Data.} To counter class imbalance and increase dataset size we follow \citet{zhang2016salient} in pretraining our model with synthetic images and gradually adding images on the SOS dataset. The images are synthesized by cut-pasting objects from the THUS10000 dataset \citep{cheng2015global} onto backgrounds from the SUN dataset \cite{xiao2010sun}. Following \cite{zhang2016salient} we apply random image transforms to each object to increase diversity in appearance. For example images we refer to Fig.~7 in \cite{zhang2016salient}.


\subsection{Implementation Details}
\label{subsec:implementation-details}
                                         
\noindent \textbf{Network Architecture.} Our models are implemented in PyTorch. \cite{paszke2017automatic}. The VAE's encoder and decoder are parameterized as CNNs. Denoting a convolutional layer as $N \,@\, F_w \times K_h - S$ with $N$ filters of size $K_w \times K_h$ and stride $S$, the encoder architecture is as follows: $64 \,@\, 4\times4-1 \rightarrow 64 \,@\, 4 \times4-2 \rightarrow 128 \,@\, 4\times4-2 \rightarrow 768 \,@\, 4\times4-2$. Final spatial features are fed in two fully-connected layers encoding the $\mu$ and $\mathrm{\Sigma}$ parameters of the posterior distribution for sampling latent vectors using the reparametrization trick \cite{kingma2013auto}. The $\mathrm{\Sigma}$ layer uses softplus activation to ensure a positive output. The decoder uses transposed convolutions \cite{zeiler2010deconvolutional} to upsample latent representations and is implemented by the following architecture: $768 \,@\, 3\times3-1 \rightarrow 768 \,@\, 3 \times3-2 \rightarrow 256 \,@\, 3\times3-2 \rightarrow 64 \,@\, 3\times3-2$. All convolutional blocks are followed by Leaky ReLU activation and batch normalization \cite{ioffe2015batch}. The feature perceptual loss parameters are set to $\alpha=1.0$ and $\beta=0.03$. The size of the latent dimension is set to $180$. \\

\noindent \textbf{Optimization.} Data augmentation of random horizontal flips, crops and color shifts is applied to all images. Preprocessed images are of $161\times161$ size when fed into the network. We warm-up by pretraining on $80$K synthetic images and gradually start adding natural images beyond $20$ epochs. The initial learning rate is set to $0.0015$ and is divided by $5$ when the loss on the test set plateaus for more than $4$ epochs. The VAE is trained for a total of $140$ epochs. To remedy the class imbalance we follow \citep{JMLR:v18:16-365} by randomly removing $10\%$ examples from the most frequent classes (and loss weighting for the softmax classifier in Sec.~\ref{subsec:unsupervised-numerosity-representation}).  

\subsection{Evaluating the Visual Numerosity Representation}
\label{subsec:unsupervised-numerosity-representation}

In this experiment we evaluate the strength of the representations learned by the VAE in unsupervised setting. As the visual reconstructions are hard to compare and do not conceal the sense visual numerosity, we perform a quantitative comparison with the state-of-the-art \cite{zhang2016salient,torralba2003context,chatfield2011devil} by training a simple softmax classifier on top of the visual representations learned without supervision. The task is to predict the instance count (classification). Specifically, we fix the VAE parameters and feed the latent representations for a given image to the softmax classifier referred to as \textbf{VAE + softmax}\footnote{The softmax classifier is modeled as two-layer multi-layer perceptron with 160 units per layer and ReLU activations. It takes as input the VAEs latent representation and predicts the instance count.}. We use the SOS train set \emph{with} count labels to minimize cross-entropy loss. Note that this count classifier uses supervision on top of the unsupervised visual representations. We compare the subitizing performance with existing work: handcrafted GIST features with SVM classifier \cite{torralba2003context}, a SIFT-based representation with Improved Fisher Vectors (IFV) \cite{chatfield2011devil} and the fully-supervised CNN specifically designed for subitizing \cite{zhang2016salient}.  

\autoref{tab:sub-performance} reports the performance of both our method and existing work on the SOS test set \cite{zhang2016salient}. The subitizing performance of our softmax classifier is comparable to the performance of the GIFT and SVM classifier \cite{torralba2003context} and SIFT+IFV \cite{chatfield2011devil}. Our method is unable to surpass the fully-supervised CNN of \cite{zhang2016salient}. This is not surprising as their network is significantly larger, pretrained on millions of images from ImageNet and uses full supervision from the SOS count labels. Our visual representations are trained in unsupervised setting without ImageNet pretraining. The quantitative results indicate that unsupervised learning of a representation of visually complex images by a VAE discovered that numerosity in the subitizing range is a key characteristic of natural scenes. 

\begin{table}[t]
\centering
\caption{Comparison of our unsupervised approach to existing supervised approaches for \emph{instance counting} over the Salient Object Subitizing dataset. We report the count average precision ($\%$) over the entire test set. Results from existing methods (rows 2-4) were reported in \cite{zhang2016salient}. Our method is unable to outperform the fully-supervised CNN of \cite{zhang2016salient} but performs well given that our visual representations are trained unsupervised. \label{tab:sub-performance}}
\vspace{1em}
    \begin{tabular*}{\textwidth}{@{\extracolsep{\fill}}lllllll}
    \toprule
    Count Label $\rightarrow$ & \quad 0 & \quad 1 & \quad 2 & \quad 3 & \quad 4+ & \quad mean \\
    \midrule
    Chance & \quad 27.5 & \quad46.5 & \quad18.6 & \quad11.7 & \quad\phantom{0}9.7 & \quad22.8 \\
    GIST \cite{torralba2003context} & \quad 67.4 & \quad65.0 & \quad32.3 & \quad17.5 & \quad24.7 & \quad41.4 \\
    SIFT+IFV \cite{chatfield2011devil} & \quad 83.0 & \quad68.1 & \quad35.1 & \quad26.6 & \quad38.1 & \quad50.1 \\
    CNN\_FT \cite{zhang2016salient} & \quad 93.6 & \quad93.8 & \quad75.2 & \quad58.6 & \quad71.6 & \quad78.6 \\
    VAE + softmax (ours) & \quad76.0 & \quad49.0 & \quad40.0 & \quad27.0 & \quad30.0 & \quad44.4 \\
    \bottomrule
    \end{tabular*}
\end{table}

\subsection{Size-invariant Numerosity Detectors}
\label{subsec:qualitative-analysis}

\begin{figure}
    \centering
    \includegraphics[width=1.0\textwidth]{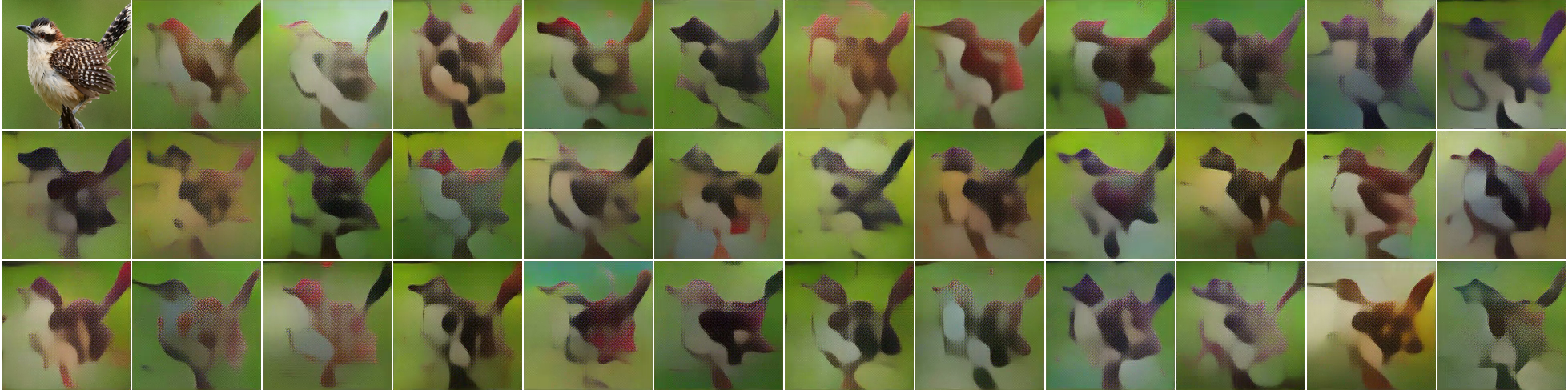}
    \caption{\textbf{Top Left:} original image with a single salient instance from SOS. \textbf{Remaining Images:} reconstructions of the VAE by slightly increasing the response at individual dimensions in the latent representation. A single value in the latent space can correspond to multiple image characteristics such as lighting, color, numerosity and object size.}
    \label{fig:latent}
\end{figure}

We continue to investigate whether the learned visual representations for numerosity estimation are invariant to \emph{object area}, as previously found in \cite{stoianov2012} for simple synthesized data and observed in cognitive studies \cite{nieder2016neuronal, harvey2013topographic}. Our methodology is similar to \citet{stoianov2012}: we attempt to determine the relationship between the VAE's latent representations and cumulative object area and instance count in synthesized images.

To this end, we create a dataset with synthetic images containing $N \in [0, 4]$ copies of the same object (with $N$ sampled uniformly at random) and their corresponding cumulative area values $A$ (measured in the number of pixels). We fix the VAE parameters and create a set of latent vectors $Z$ from $35$K synthetic images with $15$ object classes from the THUS10000 dataset \cite{cheng2015global}; the size between objects in each image varies modestly. To reduce noise in the representations created by the VAE we make sure objects are not overlapping (interestingly, overlapping objects hinder our brain's ability to subitize \cite{dehaene2011number}). Although not natural images, their appearance is more diverse than the binary images of \cite{stoianov2012}.

Using all latent vectors $z$ from $Z$, we search for latent dimensions $z_i$ that serve as either numerosity or area encoders by means of a linear regression suggested in \cite{stoianov2012}. Following their approach, we fit the following relationship between $z_i$ and the variables $A$ and $N$ across the entire dataset (all variables normalized): 
\begin{equation}
    \label{eq:regression}
    z_i = \beta_1 \log(N) + \beta_2\log(A) + \varepsilon
\end{equation}
\citet{stoianov2012} formulate two criteria for which an individual dimension $z_i$ significantly responds to changes in object size or numerosity. First, the dimension should explain at least $10\%$ of the variance ($R^2 \geq 0.1$) in the activity, and secondly, the regression coefficient of the complementary property has an absolute value smaller than $0.1$. Here, we slightly loosen the first criteria of \cite{stoianov2012} by setting the threshold to $5\%$ variance because the visual complexity of our training images is significantly higher. More specifically, in our VAE the individual latent dimensions $z_i$ can be responsible for encoding more than one visual characteristic. This is observed in \autoref{fig:latent}, where a slight change in the latent dimension can change more than one visual characteristic in the reconstruction. Due to this fact, the responses of $z_i$ will inherently be noisier. 

The results of our fit after regressing on the $35$K synthetic images are as follows. We found one or two reoccurring dimensions that responded to area $A$ or numerosity $N$ with regression error $R^2 > 0.06 \pm 0.020$.  Interestingly, we also found that whenever the regression showed multiple dimensions responding to object area, the two dimensions always changed in opposite direction (different sign) which is in agreement with \cite{stoianov2012}. Therefore, the latent space likely encodes object area and numerosity as independent properties of images, consistent with coding properties of numerosity-selective neurons \cite{harvey2013topographic}.

\begin{figure}[t]
    \centering
    \begin{subfigure}{.5\textwidth}
      \centering
      \includegraphics[scale=0.115]{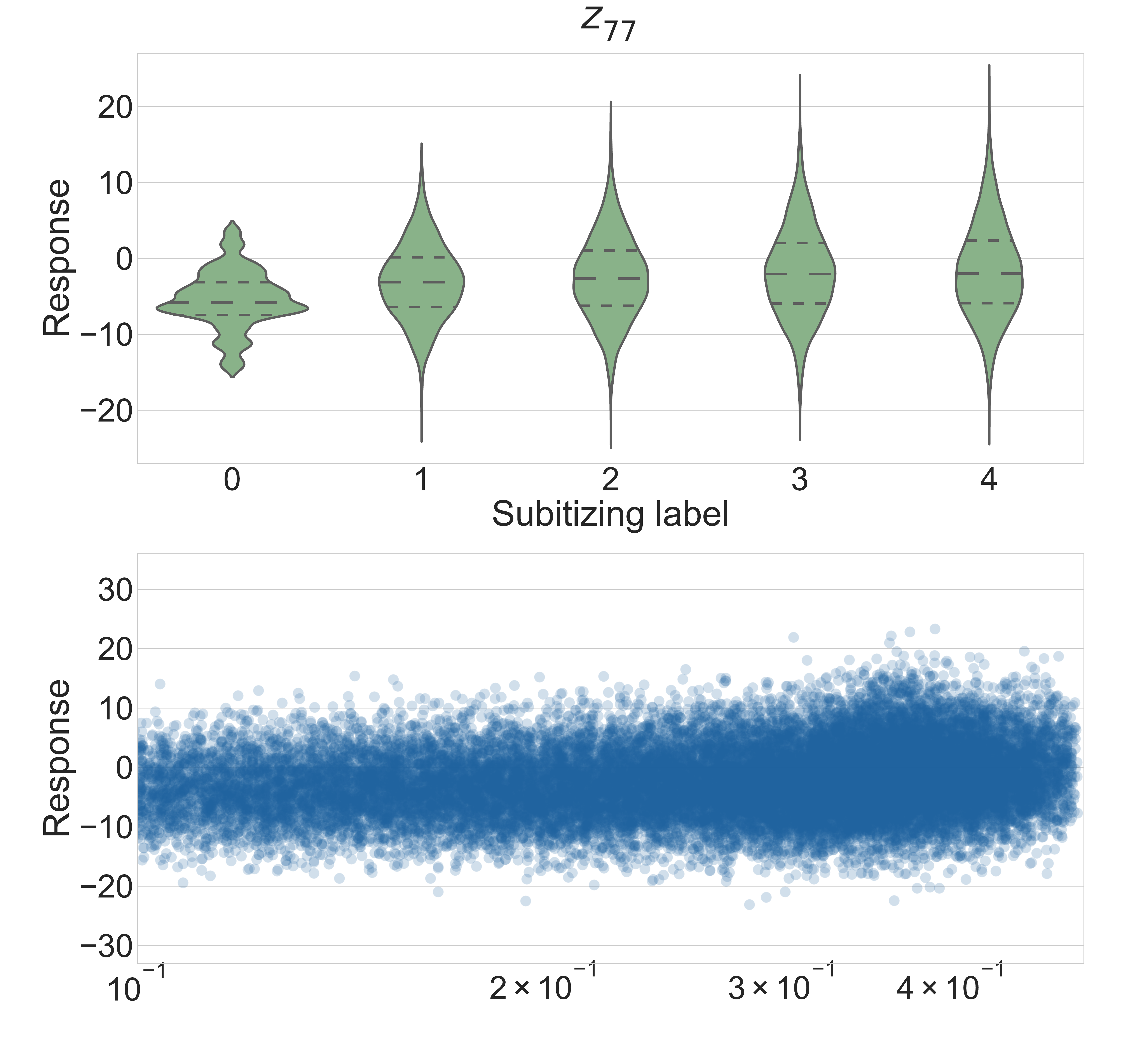}
      \caption{}
      \label{fig:zn}
    \end{subfigure}%
    \begin{subfigure}{.5\textwidth}
      \centering
      \includegraphics[scale=0.115]{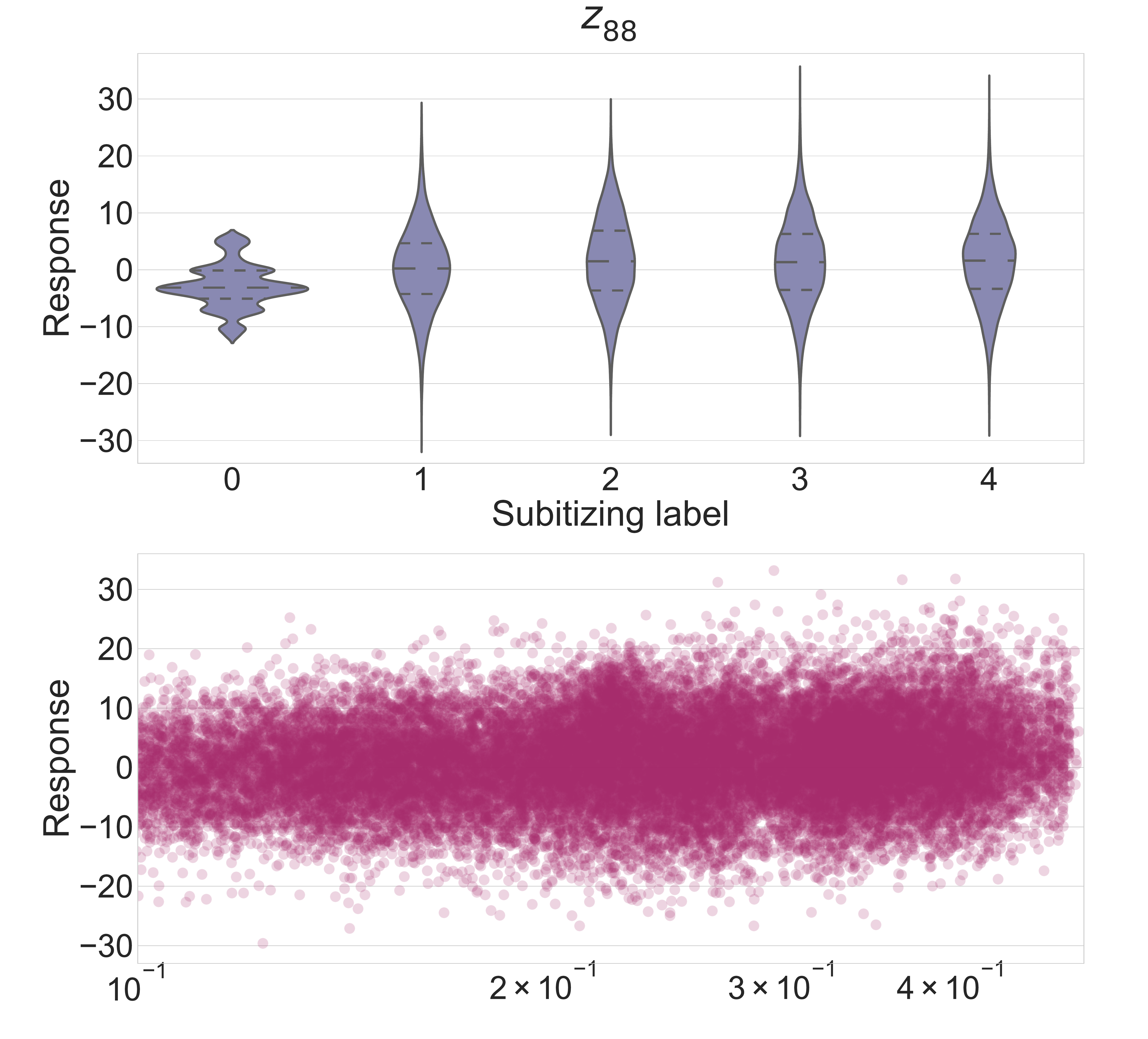}
       \caption{}
      \label{fig:za}
    \end{subfigure}
    \caption{Response distribution of two latent dimensions when feeding synthetic images of different subitizing label and cumulative object area. \textbf{(a)} $z_{77}$ responds to numerosity (subitizing label) whereas being invariant to object size ($R=0.055$ for the fit of Eq.~\eqref{eq:regression}). \textbf{(b)} $z_{88}$ shows a typical response profile of a dimension sensitive to cumulative object count ($R=0.056$). The cumulative object area $A$ is shown on logarithmic scale.}
    \label{fig:linear}
\end{figure}

Finally, in \autoref{fig:linear} we plot the characteristics of response profiles of the dimensions $z_i$ that were found to encode either cumulative area or visual numerosity. For the area dimension ($z_{88}$ shown in \autoref{fig:za}), images with either very small or large cumulative area push the mean response distribution significantly upward or downward. On the other hand, for the numerosity encoding dimension ($z_{77}$ shown in \autoref{fig:zn}) the response is more stable. This is evidence for dimension $z_{77}$ encoding visual number sense while being invariant to object size. 

\section{Conclusion}
\label{conclusion}

We have proposed unsupervised representation learning for visual number sense on natural images. Specifically, we propose a convolutional variational autoencoder to learn the concept of number from both synthetic and natural images without supervision. In agreement with previous findings on numerosity in artificial multi-layer perceptrons \citep{stoianov2012} and biological neuronal populations \citep{harvey2013topographic, nieder2016neuronal}, a representation with the ability to encode numerosity within the subitizing range invariant to object area and appearance has been learned. Therefore, we present additional evidence that the concept of visual number sense emerges as a statistical property in variational autoencoders when presented a set of images displaying a varying number of salient objects.

\bibliography{ms}

\begin{thebibliography}{45}
\providecommand{\natexlab}[1]{#1}
\providecommand{\url}[1]{\texttt{#1}}
\expandafter\ifx\csname urlstyle\endcsname\relax
  \providecommand{\doi}[1]{doi: #1}\else
  \providecommand{\doi}{doi: \begingroup \urlstyle{rm}\Url}\fi

\bibitem[Arteta et~al.(2016)Arteta, Lempitsky, and
  Zisserman]{arteta2016counting}
C.~Arteta, V.~Lempitsky, and A.~Zisserman.
\newblock Counting in the wild.
\newblock In \emph{ECCV}, 2016.

\bibitem[Burr and Ross(2008)]{burr2008visual}
D.~Burr and J.~Ross.
\newblock A visual sense of number.
\newblock \emph{Current biology}, 18\penalty0 (6):\penalty0 425--428, 2008.

\bibitem[Chatfield et~al.(2011)Chatfield, Lempitsky, Vedaldi, and
  Zisserman]{chatfield2011devil}
K.~Chatfield, V.~S. Lempitsky, A.~Vedaldi, and A.~Zisserman.
\newblock The devil is in the details: an evaluation of recent feature encoding
  methods.
\newblock In \emph{BMVC}, 2011.

\bibitem[Chattopadhyay et~al.(2017)Chattopadhyay, Vedantam, Selvaraju, Batra,
  and Parikh]{Chattopadhyay_2017_CVPR}
P.~Chattopadhyay, R.~Vedantam, R.~R. Selvaraju, D.~Batra, and D.~Parikh.
\newblock Counting everyday objects in everyday scenes.
\newblock In \emph{CVPR}, 2017.

\bibitem[Cheng et~al.(2015)Cheng, Mitra, Huang, Torr, and Hu]{cheng2015global}
M.-M. Cheng, N.~J. Mitra, X.~Huang, P.~H. Torr, and S.-M. Hu.
\newblock Global contrast based salient region detection.
\newblock \emph{PAMI}, 37\penalty0 (3):\penalty0 569--582, 2015.

\bibitem[Davis and P{\'e}russe(1988)]{animalsnumericalcognition}
H.~Davis and R.~P{\'e}russe.
\newblock Numerical competence in animals: Definitional issues, current
  evidence, and a new research agenda.
\newblock \emph{Behavioral and Brain Sciences}, 11\penalty0 (4):\penalty0
  561--579, 1988.

\bibitem[Dehaene(2011)]{dehaene2011number}
S.~Dehaene.
\newblock \emph{The number sense: How the mind creates mathematics}.
\newblock OUP USA, 2011.

\bibitem[Doersch(2016)]{doersch2016tutorial}
C.~Doersch.
\newblock Tutorial on variational autoencoders.
\newblock \emph{arXiv preprint arXiv:1606.05908}, 2016.

\bibitem[Feigenson et~al.(2004)Feigenson, Dehaene, and
  Spelke]{feigenson2004core}
L.~Feigenson, S.~Dehaene, and E.~Spelke.
\newblock Core systems of number.
\newblock \emph{Trends in cognitive sciences}, 8\penalty0 (7):\penalty0
  307--314, 2004.

\bibitem[Franka et~al.(2008)Franka, Everettb, Fedorenkoa, and
  Gibsona]{franka2008number}
M.~C. Franka, D.~L. Everettb, E.~Fedorenkoa, and E.~Gibsona.
\newblock Number as a cognitive technology: Evidence from pirah{\~a} language
  and cognitionq.
\newblock \emph{Cognition}, 108:\penalty0 819--824, 2008.

\bibitem[Goodfellow et~al.(2016)Goodfellow, Bengio, Courville, and
  Bengio]{goodfellow2016deep}
I.~Goodfellow, Y.~Bengio, A.~Courville, and Y.~Bengio.
\newblock \emph{Deep learning}, volume~1.
\newblock MIT press Cambridge, 2016.

\bibitem[Harvey et~al.(2013)Harvey, Klein, Petridou, and
  Dumoulin]{harvey2013topographic}
B.~M. Harvey, B.~P. Klein, N.~Petridou, and S.~O. Dumoulin.
\newblock Topographic representation of numerosity in the human parietal
  cortex.
\newblock \emph{Science}, 341\penalty0 (6150):\penalty0 1123--1126, 2013.

\bibitem[He et~al.(2017)He, Jiao, Zhang, Han, and Lau]{he2017delving}
S.~He, J.~Jiao, X.~Zhang, G.~Han, and R.~W. Lau.
\newblock Delving into salient object subitizing and detection.
\newblock In \emph{ICCV}, 2017.

\bibitem[Hou et~al.(2017)Hou, Shen, Sun, and Qiu]{hou2017deep}
X.~Hou, L.~Shen, K.~Sun, and G.~Qiu.
\newblock Deep feature consistent variational autoencoder.
\newblock In \emph{WACV}, 2017.

\bibitem[Hu et~al.(2016)Hu, Chang, Nian, Wang, and Li]{hu2016dense}
Y.~Hu, H.~Chang, F.~Nian, Y.~Wang, and T.~Li.
\newblock Dense crowd counting from still images with convolutional neural
  networks.
\newblock \emph{Journal of Visual Communication and Image Representation},
  38:\penalty0 530--539, 2016.

\bibitem[Ioffe and Szegedy(2015)]{ioffe2015batch}
S.~Ioffe and C.~Szegedy.
\newblock Batch normalization: Accelerating deep network training by reducing
  internal covariate shift.
\newblock In \emph{ICML}, 2015.

\bibitem[Jansen et~al.(2014)Jansen, Hofman, Straatemeier, Bers, Raijmakers, and
  Maas]{jansen2014role}
B.~R. Jansen, A.~D. Hofman, M.~Straatemeier, B.~M. Bers, M.~E. Raijmakers, and
  H.~L. Maas.
\newblock The role of pattern recognition in children's exact enumeration of
  small numbers.
\newblock \emph{British Journal of Developmental Psychology}, 32\penalty0
  (2):\penalty0 178--194, 2014.

\bibitem[Jevons(1871)]{jevons1871power}
W.~S. Jevons.
\newblock The power of numerical discrimination, 1871.

\bibitem[Kaufman et~al.(1949)Kaufman, Lord, Reese, and
  Volkmann]{kaufman1949discrimination}
E.~L. Kaufman, M.~W. Lord, T.~W. Reese, and J.~Volkmann.
\newblock The discrimination of visual number.
\newblock \emph{The American journal of psychology}, 62\penalty0 (4):\penalty0
  498--525, 1949.

\bibitem[Kingma and Welling(2014)]{kingma2013auto}
D.~P. Kingma and M.~Welling.
\newblock Auto-encoding variational bayes.
\newblock In \emph{ICLR}, 2014.

\bibitem[Lakoff and N{\'u}{\~n}ez(2000)]{lakoff}
G.~Lakoff and R.~E. N{\'u}{\~n}ez.
\newblock Where mathematics comes from: How the embodied mind brings
  mathematics into being.
\newblock \emph{AMC}, 10:\penalty0 12, 2000.

\bibitem[LeCun et~al.(2015)LeCun, Bengio, and Hinton]{lecun2015deep}
Y.~LeCun, Y.~Bengio, and G.~Hinton.
\newblock Deep learning.
\newblock \emph{Nature}, 521\penalty0 (7553):\penalty0 436, 2015.

\bibitem[Lema{{\^i}}tre et~al.(2017)Lema{{\^i}}tre, Nogueira, and
  Aridas]{JMLR:v18:16-365}
G.~Lema{{\^i}}tre, F.~Nogueira, and C.~K. Aridas.
\newblock Imbalanced-learn: A python toolbox to tackle the curse of imbalanced
  datasets in machine learning.
\newblock \emph{JMLR}, 18\penalty0 (17):\penalty0 1--5, 2017.

\bibitem[Lempitsky and Zisserman(2010)]{lempitsky2010learning}
V.~Lempitsky and A.~Zisserman.
\newblock Learning to count objects in images.
\newblock In \emph{NIPS}, 2010.

\bibitem[Levy and Wolf(2015)]{levy2015live}
O.~Levy and L.~Wolf.
\newblock Live repetition counting.
\newblock In \emph{ICCV}, 2015.

\bibitem[Lin et~al.(2014)Lin, Maire, Belongie, Hays, Perona, Ramanan,
  Doll{\'a}r, and Zitnick]{lin2014microsoft}
T.-Y. Lin, M.~Maire, S.~Belongie, J.~Hays, P.~Perona, D.~Ramanan,
  P.~Doll{\'a}r, and C.~L. Zitnick.
\newblock Microsoft coco: Common objects in context.
\newblock In \emph{ECCV}, 2014.

\bibitem[Liu et~al.(2016)Liu, Wang, Feng, and Xi]{liu2016highway}
X.~Liu, Z.~Wang, J.~Feng, and H.~Xi.
\newblock Highway vehicle counting in compressed domain.
\newblock In \emph{CVPR}, 2016.

\bibitem[Nieder(2016)]{nieder2016neuronal}
A.~Nieder.
\newblock The neuronal code for number.
\newblock \emph{Nature Reviews Neuroscience}, 17\penalty0 (6):\penalty0
  366--382, 2016.

\bibitem[Nieder and Dehaene(2009)]{nieder2009representation}
A.~Nieder and S.~Dehaene.
\newblock Representation of number in the brain.
\newblock \emph{Annual review of neuroscience}, 32:\penalty0 185--208, 2009.

\bibitem[Noroozi et~al.(2017)Noroozi, Pirsiavash, and
  Favaro]{noroozi2017representation}
M.~Noroozi, H.~Pirsiavash, and P.~Favaro.
\newblock Representation learning by learning to count.
\newblock In \emph{ICCV}, 2017.

\bibitem[Paszke et~al.(2017)Paszke, Gross, Chintala, Chanan, Yang, DeVito, Lin,
  Desmaison, Antiga, and Lerer]{paszke2017automatic}
A.~Paszke, S.~Gross, S.~Chintala, G.~Chanan, E.~Yang, Z.~DeVito, Z.~Lin,
  A.~Desmaison, L.~Antiga, and A.~Lerer.
\newblock Automatic differentiation in pytorch.
\newblock In \emph{NIPS Workshops}, 2017.

\bibitem[Piazza and Izard(2009)]{piazza2009humans}
M.~Piazza and V.~Izard.
\newblock How humans count: numerosity and the parietal cortex.
\newblock \emph{The {N}euroscientist}, 15\penalty0 (3):\penalty0 261--273,
  2009.

\bibitem[Poncet et~al.(2016)Poncet, Caramazza, and
  Mazza]{poncet2016individuation}
M.~Poncet, A.~Caramazza, and V.~Mazza.
\newblock Individuation of objects and object parts rely on the same neuronal
  mechanism.
\newblock \emph{Scientific reports}, 6:\penalty0 38434, 2016.

\bibitem[Runia et~al.(2018)Runia, Snoek, and Smeulders]{runia2018repetition}
T.~F.~H. Runia, C.~G.~M. Snoek, and A.~W.~M. Smeulders.
\newblock Real-world repetition estimation by div, grad and curl.
\newblock In \emph{CVPR}, June 2018.

\bibitem[Russakovsky et~al.(2015)Russakovsky, Deng, Su, Krause, Satheesh, Ma,
  Huang, Karpathy, Khosla, Bernstein, et~al.]{russakovsky2015imagenet}
O.~Russakovsky, J.~Deng, H.~Su, J.~Krause, S.~Satheesh, S.~Ma, Z.~Huang,
  A.~Karpathy, A.~Khosla, M.~Bernstein, et~al.
\newblock Imagenet large scale visual recognition challenge.
\newblock \emph{IJCV}, 115\penalty0 (3):\penalty0 211--252, 2015.

\bibitem[Segu{\'\i} et~al.(2015)Segu{\'\i}, Pujol, and
  Vitria]{segui2015learning}
S.~Segu{\'\i}, O.~Pujol, and J.~Vitria.
\newblock Learning to count with deep object features.
\newblock In \emph{CVPR Workshops}, 2015.

\bibitem[Simonyan and Zisserman(2015)]{simonyan2014very}
K.~Simonyan and A.~Zisserman.
\newblock Very deep convolutional networks for large-scale image recognition.
\newblock In \emph{ICLR}, 2015.

\bibitem[Stoianov and Zorzi(2012)]{stoianov2012}
I.~Stoianov and M.~Zorzi.
\newblock Emergence of a ‘visual number sense’ in hierarchical generative
  models.
\newblock \emph{Nature Neuroscience}, 15\penalty0 (2):\penalty0 194, 2012.

\bibitem[Torralba et~al.(2003)Torralba, Murphy, Freeman, Rubin,
  et~al.]{torralba2003context}
A.~Torralba, K.~P. Murphy, W.~T. Freeman, M.~A. Rubin, et~al.
\newblock Context-based vision system for place and object recognition.
\newblock In \emph{ICCV}, 2003.

\bibitem[Viswanathan and Nieder(2013)]{viswanathan2013neuronal}
P.~Viswanathan and A.~Nieder.
\newblock Neuronal correlates of a visual “sense of number” in primate
  parietal and prefrontal cortices.
\newblock \emph{Proceedings of the National Academy of Sciences}, 110\penalty0
  (27):\penalty0 11187--11192, 2013.

\bibitem[Walach and Wolf(2016)]{walach2016learning}
E.~Walach and L.~Wolf.
\newblock Learning to count with cnn boosting.
\newblock In \emph{ECCV}, 2016.

\bibitem[Xiao et~al.(2010)Xiao, Hays, Ehinger, Oliva, and
  Torralba]{xiao2010sun}
J.~Xiao, J.~Hays, K.~A. Ehinger, A.~Oliva, and A.~Torralba.
\newblock Sun database: Large-scale scene recognition from abbey to zoo.
\newblock In \emph{CVPR}, 2010.

\bibitem[Xiong et~al.(2017)Xiong, Shi, and Yeung]{xiong2017spatiotemporal}
F.~Xiong, X.~Shi, and D.-Y. Yeung.
\newblock Spatiotemporal modeling for crowd counting in videos.
\newblock In \emph{ICCV}, 2017.

\bibitem[Zeiler et~al.(2010)Zeiler, Krishnan, Taylor, and
  Fergus]{zeiler2010deconvolutional}
M.~D. Zeiler, D.~Krishnan, G.~W. Taylor, and R.~Fergus.
\newblock Deconvolutional networks.
\newblock In \emph{CVPR}, 2010.

\bibitem[Zhang et~al.(2017)Zhang, Ma, Sameki, Sclaroff, Betke, Lin, Shen,
  Price, and M{\u{e}}ch]{zhang2016salient}
J.~Zhang, S.~Ma, M.~Sameki, S.~Sclaroff, M.~Betke, Z.~Lin, X.~Shen, B.~Price,
  and R.~M{\u{e}}ch.
\newblock Salient object subitizing.
\newblock \emph{IJCV}, 124\penalty0 (2):\penalty0 169--186, 2017.

\end{thebibliography}

\end{document}